\documentclass[letterpaper,10 pt,conference]{ieeeconf}  
\IEEEoverridecommandlockouts                              
\usepackage{amsmath}
\usepackage{amssymb}
\usepackage{graphicx}
\usepackage{graphics}
\usepackage{blkarray}
\usepackage{booktabs}
\usepackage{xcolor} 
\usepackage{multirow}
\usepackage{comment}
\usepackage{epstopdf}
\usepackage{tikz}
\usetikzlibrary{arrows,matrix,positioning}
\usepackage{todonotes}
\usepackage[labelformat=simple]{subcaption}
\usepackage{soul}
\usepackage{MnSymbol}
\usepackage{cite}
\usepackage{wrapfig}
\usepackage{units}
\usepackage{xfrac}
\usepackage{algpseudocode}
\usepackage{algorithm}

\graphicspath{{./figures/}}

\usepackage{tikz}
\newcommand\copyrighttext{%
  \footnotesize Accepted for presentation in IROS 2016, Daejeon, Korea. \copyright 2016 IEEE. Personal use of this material is permitted. Permission from IEEE must be obtained for all other uses, in any current or future media, including reprinting/republishing this material for advertising or promotional purposes, creating new collective works, for resale or redistribution to servers or lists, or reuse of any copyrighted component of this work in other works.}
\newcommand\copyrightnotice{%
\begin{tikzpicture}[remember picture,overlay]
\node[anchor=south,yshift=10pt] at (current page.south) {\fbox{\parbox{\dimexpr\textwidth-\fboxsep-\fboxrule\relax}{\copyrighttext}}};
\end{tikzpicture}%
}

\title{\LARGE \bf
Active Exploration Using Gaussian Random Fields \linebreak and Gaussian Process Implicit Surfaces
}
\author{ Sergio Caccamo, Yasemin Bekiroglu, Carl Henrik Ek, Danica Kragic
\thanks{Caccamo, Ek and Kragic are with the Computer Vision and Active Perception Lab., Centre for Autonomous Systems, School of Computer Science and Communication, Royal Institute of Technology (KTH), SE-100 44 Stockholm, SE.
e-mail: \tt{ $\{$caccamo$|$chek$|$dani$\}$@kth.se}}
\thanks{Bekiroglu is with the School of Mechanical Engineering, University of Birmingham, UK
e-mail: \tt{ $\{$Y.Bekiroglu$\}$@bham.ac.uk}}}

\begin{document}
\maketitle
\thispagestyle{empty}
\pagestyle{empty}
\begin{abstract}

In this work we study the problem of exploring surfaces and building compact 3D representations of the environment surrounding a robot through active perception.
We propose an online probabilistic framework that merges visual and tactile measurements using Gaussian Random Field and Gaussian Process Implicit Surfaces.
The system investigates incomplete point clouds in order to find a small set of regions of interest which are then physically explored with a robotic arm equipped with tactile sensors.
We show experimental results obtained using a PrimeSense camera, a Kinova Jaco2 robotic arm and Optoforce sensors on different scenarios. We then demonstrate how to use the online framework for object detection and terrain classification.

\end{abstract}

 \copyrightnotice

\section{Introduction}
\label{sec:intro}

\begin{figure}[t]
  \center
     \includegraphics[width=\columnwidth]{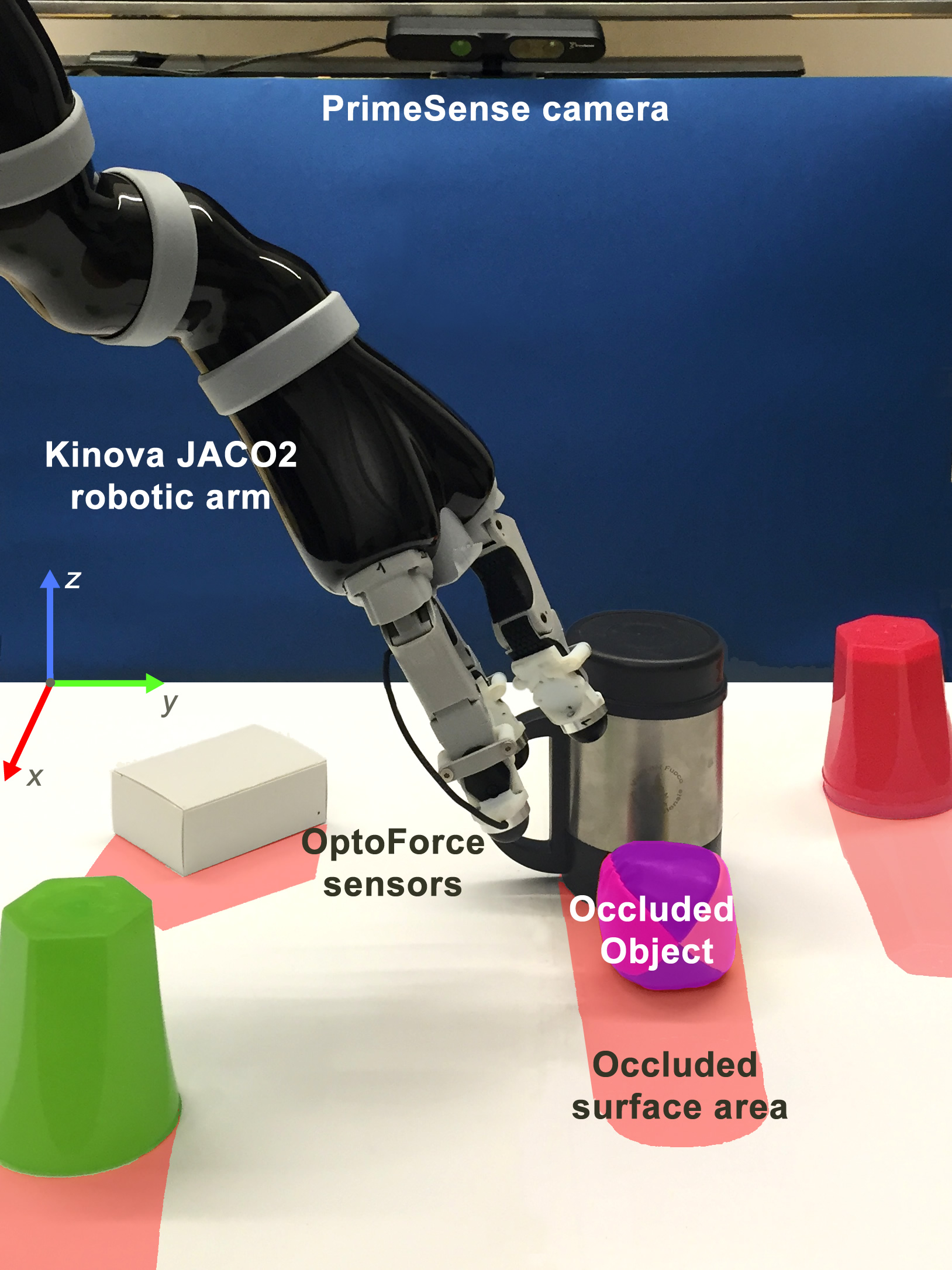}
  \caption{\small{Experimental setup. A kinova Jaco2 arm equipped with Optoforce sensor explores occluded areas of an environment. }}
  \label{fig:setup_arm}
\end{figure}

Acquiring a high quality 3D model of the environment is critical for many autonomous robotics problems such as grasping, segmentation, traversability or mapping.
Mere vision perception does not often exhaustively describe the shape of the environment since volumetric data generated from modern vision sensors are prone to errors due to limited field of view, photometric effects, occlusions and noise.

Passive observation of a scene leads to incomplete shapes of objects and terrains facing the camera. Heuristic or symmetry assumptions \cite{bohggap} can be used to deal with lack of data in the observations, leading to errors in the representation.

Surface exploration through vision and haptic interactions is the task of purposefully touch and inspect a portion of environment so to reveal occluded information. It can be considered as a case of either interactive or active perception depending whether the physical interaction strategically modify the environment under analysis or not respectively.

Haptic exploration helps improving observations adding a new layer of information into the world model. Meier et al.\cite{meierprob} showed that tactile information alone can be used to adequately describe objects properties.
A robotic manipulator equipped with tactile sensors can be used to encode properties of surfaces and objects and enhance visual perception of shapes\cite{yasemingp}.  Studies \cite{helbigperc} show that combining tactile and visual representations of object brings more reliable and robust shape estimation than either the visual or tactile alone. Even humans build their world representation using different sensory information and actively examine the environment to enhance their perception \cite{ernstperc}.

\begin{figure}[t]
  \center
     \includegraphics[width=\columnwidth]{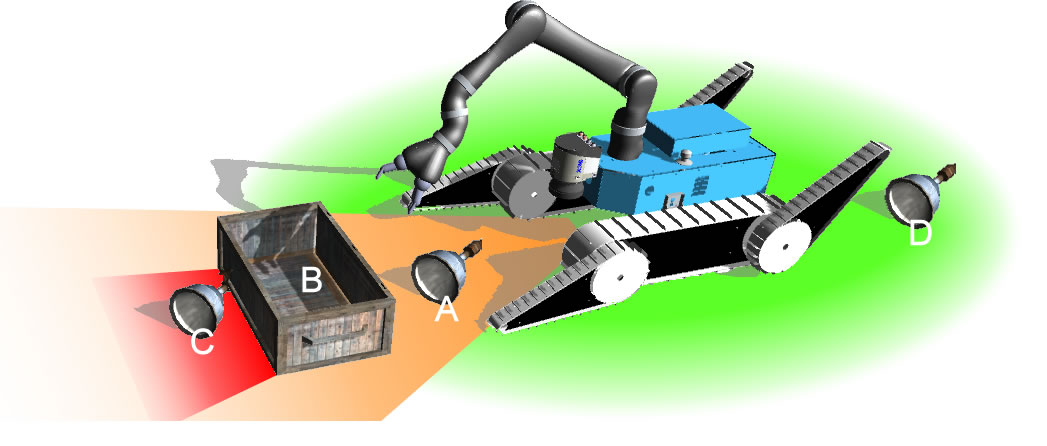}
  \caption{\small{Visible and reachable surface of an UGV. Objects A and B are inside the VRS, B is partially occluded. C and D are outside the VRS. }}
  \label{fig:vrs_pic}
\end{figure}

In this paper we use the term VRS (visible and reachable surface) to refer to all the surfaces contained in the portion of space consisting in the intersection of the field of view of the camera and the reachable space of the robot.
The VRS can contain several occluded regions and objects (i.e. occluded surfaces, see Fig. \ref{fig:setup_arm}). \\
Examples of VRS are:
\begin{itemize}
\item[-] The surface of a table and objects placed in front of a robot.
\item[-] The portion of map in front of an arm equipped Unmanned Ground Vehicle (UGV) as shown in Fig. \ref{fig:vrs_pic}.
\end{itemize}

Inspired by the work in \cite{yasemingp} and field applications described in \cite{tradr} we build 3D models of a VRS by merging vision and haptic data into a probabilistic framework. We study how to properly model shape of environments that contain different occluded objects and incomplete areas taking advantage of uncertainties in the sensory system. We also show how to reduce the exploration time and the amount of physical interactions needed.

In summary, the contributions of this paper can be listed as follow: 
\begin{enumerate}
\item We propose a new probabilistic framework for surface exploration that merges haptic and visual sensory information for building a local 3D map of the environment.
\item We show how to exploit Gaussian Processes and Delaunay triangulation for reducing the amount of interactions and the computational power/time needed.
\item We demonstrate the feasibility of the approach through hardware experiments letting a robotic arm explore different scenarios and show how few interactions can add useful information to a partially visible surface.
\item We show how to generically exploit the autonomous framework for problems such as object identification and terrain classification.
\end{enumerate}

\section{Related Work}
\label{sec:relatedWork}

Gaussian processes (GP) \cite{rasmussengp} have been used for terrain mapping and modeling \cite{vasudevangpt}, \cite{callaghangpt} in a wide range of applications including geophysics, aeronautics and robotics. In those studies any given point in a 2D Euclidean space is associated with a single elevation value, generating therefore a 2.5D surface called digital elevation model (DEM) also known as heightmap\cite{peckhamdem}. Another example of Gaussian processes applied to digital terrain modeling can be seen in \cite{vasudevangpls}. Such models may fit terrain shapes but are not suitable for more complex surfaces or in applications such as object manipulation or segmentation.
 
Implicit surfaces\cite{turkis} have been widely used to represent object shapes since their first appearances in \cite{blinnis}. Generic reconstruction of implicit surfaces from data points have been presented in \cite{soleunorganized}. Machine learning techniques have been progressively developed to represent complex surfaces as in \cite{steinkesvm} and \cite{ohtakemlp}. More recently, Gaussian Process Implicit Surfaces (GPIS)\cite{williamsgpis} have become very popular allowing to extend implicit surface to uncertainty, a property needed when the model is the result of sensory data fusion\cite{beltagygp}. Environmental observations can condition a GP so that its posterior mean define the implicit surface defining the terrain (including objects). Authors in \cite{marcosgpt} applied GPIS for building 3D representation of the environment by fusing laser and mm-wave radar data. Results in \cite{dragievgp} show how GPIS as object representation can even improve complex tasks as grasping.

One disadvantage of these approaches is that during inference, Gaussian Process Regression (GPR) is computationally demanding. The major cost takes place from the inversion of large covariance matrices that, in the simplest implementation, have complexity $O\left(n^3\right)$. Mathematical tools as Cholesky decomposition or sparse kernels \cite{melkumyangp} can considerably reduce the computational effort. 

GPIS requires a dense cubic matrix of points as test set in order to qualitatively describe the implicit surface. When a VRS includes several objects, the matrix containing the training sample points (i.e. the point cloud) becomes large and the computational time increases drastically making the implementation of an online active perception algorithm for surface exploration infeasible. To overcome such problem a down-sampled subset is often heuristically selected and used \cite{lawrencesgp}. 

An application of sparse kernels for mapping of large area is presented in \cite{soohwangpmap} where authors propose a unified framework for building continuous occupancy grid maps. As already mentioned, merging haptic and visual sensory data into the same probabilistic model using GPs can lead to better shape representation \cite{yasemingp} and planning \cite{hollingerui}. An example of tactile sensing for object tracking with visual occlusion using particle filters is presented in \cite{zangparticlefilt}. Another efficient tactile perception algorithm for object manipulation and tracking is shown in \cite{petrovskayapf}.

Work in \cite{zimmermann2015adaptive} and \cite{zimmermann2014adaptive} showed that the morphology of an environment is very important for an unmanned ground vehicle (UGV) that autonomously tries to traverse a particularly harsh environment. They showed also that occlusion and reflections, e.g. caused by a pot of water or broken glass, can lead to failure. This could be solved by asking a robotic arm to strategically explore the environment around the UGV. Nevertheless, the battery capacity, the computational power on board of the robot and limited time constraints, common during urban search and rescue missions, force the exploration task to reduce as much as possible the required elaboration time and the tactile interactions.
On a different scenario, an interactive humanoid robot, which explore a table with objects for segmentation, can try to identify hidden elements by touching occluded regions and move its head only in case of positive tactile feedback.

To address those problems we propose a probabilistic method that identifies and analyses occluded regions in the working space area of a robot, where a VRS point cloud can be much larger than a single object (e.g. cup or bottle).
We train a Gaussian Random Field (GRF) and a Gaussian Process Implicit Surface on the initial point cloud representing the VRS. We then infer the joint distribution of Gaussian Random Field model on regions of interest obtained from a 2.5D fast Delaunay triangulation. Delaunay triangulation on a discrete Euclidean d-dimensional point set corresponds to the dual graph of the Voronoi diagram \cite{fortunedly} for the same set. We use it to quickly identify and investigate large sparse areas in the visual point cloud that could potentially carry high uncertainty in the internal probabilistic model.
 We use a robot manipulator with tactile sensors for autonomously touching the isolated regions of the surface. We define a new training set for the GPIS using on-surface and off-surface tactile points obtained during each interaction by tactile sensors placed on the fingertips of the robotic hand. Finally we generate the new 3D shape inferring the GPIS on a subset of the VRS selected using the predicted mean of the GRF.

A surface exploration step in this paper denotes a single iteration of the algorithm that includes several physical interactions with the environment. During each exploration step the GPIS model is updated many times enlarging its training set with tactile information. We assume the environment to remain static during the whole analysis.

\section{Surface Modeling}
\label{sec:modelling}

In this section we briefly discuss Gaussian Processes Regression (GPR) \cite{rasmussengp}. We describe how to exploit two dimensional GPR (Gaussian Random Field) and three dimensional GPR (Gaussian Process Implicit Surfaces \cite{williamsgpis}) for modeling terrain and object shapes.

\subsection{Gaussian Random Fields}
\label{sec:gaussianprocessesrf}

We denote $\textit{P}_{VRS} = \{\mathbf{x_1},\mathbf{x_2}\dotsc\mathbf{x_N}\}$ with $\mathbf{x_i} \in \mathbb{R}^3$ the set of measurements of 3D points lying on the visible reachable surface and $D_{RF} = \{\mathbf{x_i},y_i \}_{i=1}^N $  a bi-dimensional  training set where $\mathbf{x_i}\in \mathbf{X}\subset\mathbb{R}^2$ are the \textit{xy}-coordinates of the points in $\textit{P}_{VRS}$ and $y_i$ the \textit{z}-coordinates (heights)\footnote{ Axis are described considering the frame represented in Fig.\ref{fig:setup_arm}}. We also define a set $\mathbf{X_*} \equiv \mathbf{X_{{rf}_*}} \subset\mathbb{R}^2$ of $M$ test points.
With a function $f:\mathbb{R}^2\to \mathbb{R}$ we map a 2.5D surface where each vector of \textit{xy}-coordinates generates a single height.
Such a function can efficiently be modeled by a GPR which places a multivariate Gaussian distribution over the space of $f\left(\mathbf{x}\right)$. The GP can be fully described by a mean function $m\left(\mathbf{x}\right)$ and a covariance function $k\left(\mathbf{x_i},\mathbf{x_j}\right)$. Assuming noisy observation $ y = f\left(\mathbf{x}\right) + \epsilon  \text{ with } \epsilon \sim \mathcal{N}\left( 0, \sigma_n^2\right)$ and  $m\left(\mathbf{x}\right) = 0$ the joint Gaussian distribution on the test set $\mathbf{X_*}$ becomes

\begin{equation}
\begin{bmatrix}
  \mathbf{y} \\
  \mathbf{f_*}
\end{bmatrix}
\sim \mathcal{N}\left( 0,
\begin{bmatrix}
  \mathbf{K} + \sigma_n^2I & \mathbf{k_*}\\
  \mathbf{k_*^T} & k_{**}
\end{bmatrix}
\right)
\label{eqn:predictivefunct}
\end{equation}

where $\mathbf{K}$ is the covariance matrix between the training points $\left[\mathbf{K}\right]_{i,j = 1 \dotsc N} = k\left(\mathbf{x_i},\mathbf{x_j}\right)$, $\mathbf{k_*}$ the covariance matrix between training and test points $\left[\mathbf{k_*}\right]_{i=1 \dotsc N,j=1 \dotsc M} = k\left(\mathbf{x_i},\mathbf{{x_*}_j}\right)$ and $\mathbf{k_{**}}$ the covariance matrix between the test points $\left[\mathbf{k_{**}}\right]_{i,j=1 \dotsc M} = k\left(\mathbf{{x_*}_i},\mathbf{{x_*}_j}\right)$.

The predictive function is obtained by conditioning on the training points
\begin{equation}
p\left(f_* | \mathbf{X_*},\mathbf{X},\mathbf{y}\right) =  \mathcal{N}\left(\overline{f_*},\mathbb{V}\left[f_*\right]\right)
\label{eqn:predictivefunctcond}
\end{equation}

\begin{equation}
\overline{f_*} = \mathbf{k_*^T}\left(\mathbf{K} + \sigma_n^2\mathbf{I}\right)^{-1}\mathbf{y}
\label{eqn:eqmeangp}
\end{equation}
\begin{equation}
V\left[f_*\right] = \mathbf{k_{**}}-\mathbf{k_*^T}\left(\mathbf{K} + \sigma_n^2\mathbf{I}\right)^{-1}\mathbf{k_*}
\label{eqn:eqsdgp}
\end{equation}

For this study we choose to use the popular squared exponential kernel 
\begin{equation}
k\left(\mathbf{x_i},\mathbf{x_j}\right) = \sigma_e^2 \text{exp}\left( - \frac{\left(\mathbf{x_i}-\mathbf{x_j}\right)^T\left(\mathbf{x_i}-\mathbf{x_j}\right)}{\sigma_w^2}\right)
\label{eqn:kernelexp}
\end{equation}

Gaussian random field (GRF) is a common way to refer to Gaussian Process Regressors that generalize over bi-dimensional Euclidean vectors.
Associating every coordinate to a single height is a big limitation when it comes to represent complex surfaces, e.g. mugs, inclined boxes. On the other hand, inferring a random field will directly produce 3D points by combining input-output into vectors of coordinates. This explicit behavior of the joint distribution permits to quickly obtain a DEM querying large portion of the VRS using only few bi dimensional testing points. The variance of the random field allows to directly highlight regions of low density data, e.g. occluded portion of the VRS,  or high complexity portion of surface, e.g. different heights for the same coordinate.

\subsection{Gaussian Processes Implicit Surfaces}
\label{sec:gaussianprocessesis}

Gaussian Process Implicit Surface (GPIS) models a function $f:\mathbb{R}^{3}\to \mathbb{R}$ with supporting points defining an implicit surface. Whereas equations \ref{eqn:predictivefunct}, \ref{eqn:predictivefunctcond}, \ref{eqn:eqmeangp}, \ref{eqn:eqsdgp} maintain the same form, $D_{IS} = \{\mathbf{x_i},y_i \}_{i=1}^N $ becomes the new training set where $\mathbf{x_i}\in \mathbf{X}\subseteq \textit{P}_{VRS} $ and $y_i \in \mathbb{R}$ defined as in \cite{williamsgpis}

\begin{equation}
 y_i  
  \begin{cases}
   = -1, 	& \quad \text{if } \mathbf{x_i} \text{ is below surface}\\
   = 0,     & \quad \text{if } \mathbf{x_i} \text{ is on the surface}\\
   = 1,  	& \quad \text{if } \mathbf{x_i} \text{ is above the surface}\\
  \end{cases}
  \label{eqn:yvalue}
\end{equation}

We also redefine the set $\mathbf{X_*}\equiv\mathbf{X_{{is}_*}}\subset\mathbb{R}^3$ of $M$ test points.
The implicit nature of the GPIS does not allow to directly shape the VRS. It is needed to define a large set of test points, e.g. a dense cubic volume centered on a region of interest, and then find the isosurface of value 0 on the $\overline{f_*}$  associated with the inferred points in $\mathbf{X_*}$. This operation is very computationally expensive depending on the size of the VRS and $\mathbf{X_*}$. On the other hand GPIS allows to model complex surfaces and to use not only points belonging on the surface to shape the GPR but also empty region points (i.e. $f(x)\neq0$ ).

As we show on Sec. \ref{sec:Experiments}, this property is critical in defining the amount of interactions needed to describe the occluded VRS.
For the GPIS we choose the same rbf covariance function as in Eq. \ref{eqn:kernelexp}.

Hyper-parameters were empirically chosen based on a set of experiments made on a $1m^3$ area.
Having a covariance function that maps the uncertainty on input data similarly for the bi-dimensional case and the three-dimensional case is a fundamental assumption for our analysis as we show later in section \ref{sec:expincomplete}.

\section{Methodology}
\label{sec:Methods}

\begin{figure*}[t!h]
  \center
     \includegraphics[width=\textwidth]{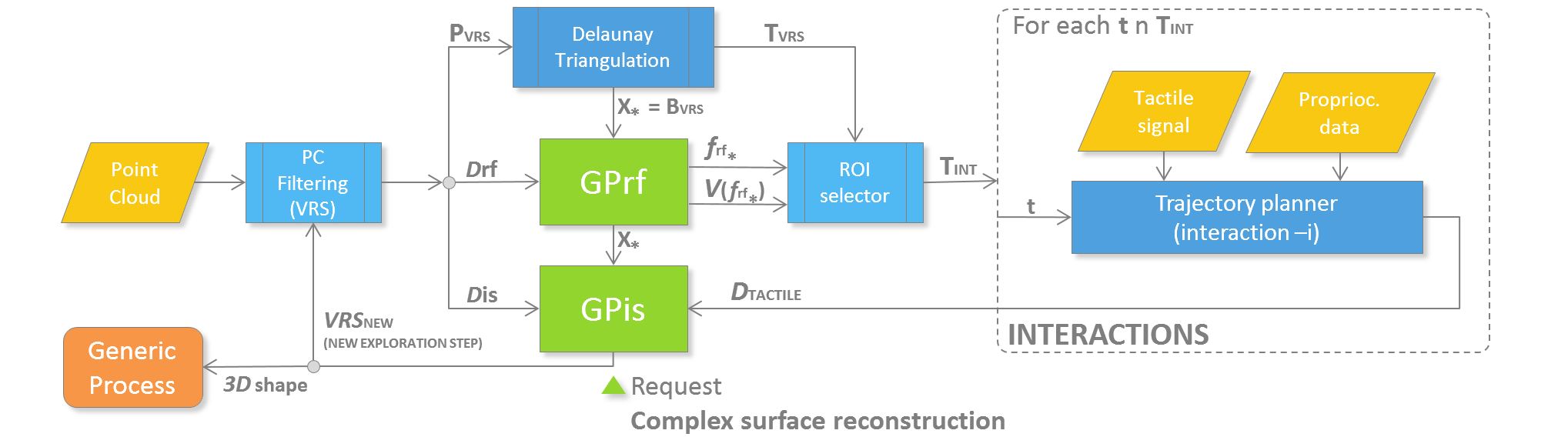}
  \caption{Probabilistic framework processes flow}
  \label{fig:process_flow}
\end{figure*}

In our study we represent the environment assuming a limited number of interactions in a constrained time window considering a VRS containing multiple objects. This applies to all those robotics problems in which a robot needs to explore an environment in an online process that extends within limited amount of time (as those listed in Sec. \ref{sec:relatedWork}). 
In this section we describe the system and guide the reader through the framework process flow showed in Fig.\ref{fig:process_flow}.

\subsection{ Strategy for modeling and inference}
\label{sec:roiselection}

We initially model the VRS surface with a 2.5D function $f_{VRS}(\mathbf{x}):\mathbb{R}^2 \to \mathbb{R} $ using a bi-dimensional Gaussian Process Regression defined in Sec. \ref{sec:gaussianprocessesrf}. 
Generalization provided by equations \ref{eqn:eqmeangp} and \ref{eqn:eqsdgp} are generally used to obtain a DEM of the VRS and the level of confidence in the data for each point. The choice of $\mathbf{X}_*$ is therefore crucial for understanding properties of the environment. A naive approach could create a dense grid of bi-dimensional points on the whole surface (so that every single hole in the map is somehow inferred by the GPR). This creates a large set $\mathbf{X}_*$ that leads to a large covariance matrix even for the bi dimensional case.
Instead of creating the grid we query only small empty regions analyzing the Voronoi diagram \cite{fortunedly} of $\textit{P}_{VRS}$ defined in \ref{sec:gaussianprocessesrf}. We run a fast 2.5D Delaunay triangulation\footnote{FADE25D C++ Library} on the set $\textit{P}_{VRS}$ and fill $\mathbf{X}_*$ with the xy-coordinates of the barycenters ($\mathbf{B_{VRS}}$) of all the computed Delaunay triangles ($\mathbf{T_{VRS}}$).
Elements in $\mathbf{X}_*$ represent coordinates of empty spaces and could easily be reduced in number, if needed, by putting a constraint on the area of the triangles to be analyzed (larger areas mean larger empty regions). 
We then use Eq. \ref{eqn:eqsdgp} to get the variance on the test points, i.e. the confidence seen as complexity of the shape or lack of information.
Computing $V\left[f_{{rf}_*}\right]$ we can select the points in $\mathbf{B_{VRS}}$ carrying highest uncertainty (ROI selector box in Fig.\ref{fig:process_flow}) and therefore also the vertices of the corresponding triangles that we denote as $\mathbf{T_{INT}} = { [\mathbf{b_1},\mathbf{tr_1}] , [\mathbf{b_2},\mathbf{tr_2}]\dotsc [\mathbf{b_L},\mathbf{tr_L}]}$ with $L<<N$.
This step is crucial for the algorithm efficiency, allowing to obtain several regions of interest in a fast way without actually inferring a Gaussian Process Implicit Surface on the whole cube containing the VRS. 

Each $\mathbf{b_i}$ is considered as a target position point for the trajectory planner for the arm . We define the approach vector for each target point (i.e. each interaction) by computing the normal vector to the plane defined by the triangle vertices constrained with direction going inside the surface (each $\mathbf{t} \in \mathbf{T_{INT}}$) .

We define an implicit surface by the support points of a function $\Psi_{VRS}(x):\mathbb{R}^3 \to \mathbb{R} $ using a Gaussian Process implicit surface  Regression defined in Sec. \ref{sec:gaussianprocessesis}. We train the model using all the 3D points in $\textit{P}_{VRS}$ labeled as 0 with the addition of a smaller\footnote{We used $\sfrac{1}{5}$N points above and $\sfrac{1}{5}$N points below the surface.} set of exterior points labeled according to Eq. \ref{eqn:yvalue}. Artificial points above and below surface are created by increasing and decreasing the z-coordinates of copies of uniformly randomly selected points in $\textit{P}_{VRS}$ respectively.

\subsection{Surface exploration}
\label{sec:surfest}

The tactile exploration task starts by sending trajectories to the robotic manipulator equipped with tactile sensors. From each sensor we obtain a temporal signal as a sequence of 3D points (sensor positions w.r.t. the world frame) along with their contact forces expressed as 3D vectors.
We define a new training set $D_{tactile} = \{\mathbf{x_i},y_i \}_{i=1}^N $ containing sampled 3D sensor positions labeled as 0 if there is contact (estimated from the module of the contact force vector) or 1 ( i.e. above the surface) if there is no contact. In case of contact we further add below-the-surface samples as virtually generated 3D points placed a few millimeters from the contact positions along the direction of the contact force (see Fig.\ref{fig:points_adding}).
We train again the GPIS model alone adding $D_{tactile}$ to the initial training set.
As we show in the experiment section on-surface points (contact points) and off-surface points are equally important when defining a surface shape. When the arm approaches the surface we start adding above-surface points to the GPIS model that will "push down" the uncertainty and redefine the implicit surface with a "clay-like" behavior.
Inside an exploration step the framework updates the internal GPIS many times and queries only the GRF. This is done so that an external concurrent process can require the last world representation available at any time and interrupt the exploration step if needed (e.g. an external process queries a small portion of space using the GPIS and identifies that the analyzed occlusion contains an object so no further interactions are required).
When all the trajectories have been used the exploration step is completed.
The inferred mean of the Gaussian random field $\overline{f_{{VRS}_*}}$ along with its variance can help reducing the dimension of the test set of the GPIS in case a full 3D model of the VRS is needed. Alg. \ref{alg:generate_dataset} shows one way to do it. It creates a test set as a grid of 2D points on the space covered by the VRS and then computes the mean function along with the variance from the GRF. It then generates 3D points for the GPIS test set using the means as xy-coordinates and the variances as confidence intervals where to span the z-coordinates. Such simple approach can help generating a test set with dimensionality considerably smaller than a dense cube of 3D points. The new VRS obtained inferring the GPIS on the new test set can be used to close the modeling loop and trigger a new iteration of the algorithm for better shape accuracy.

\begin{figure*}[t!]
  \center
     \includegraphics[width=\textwidth]{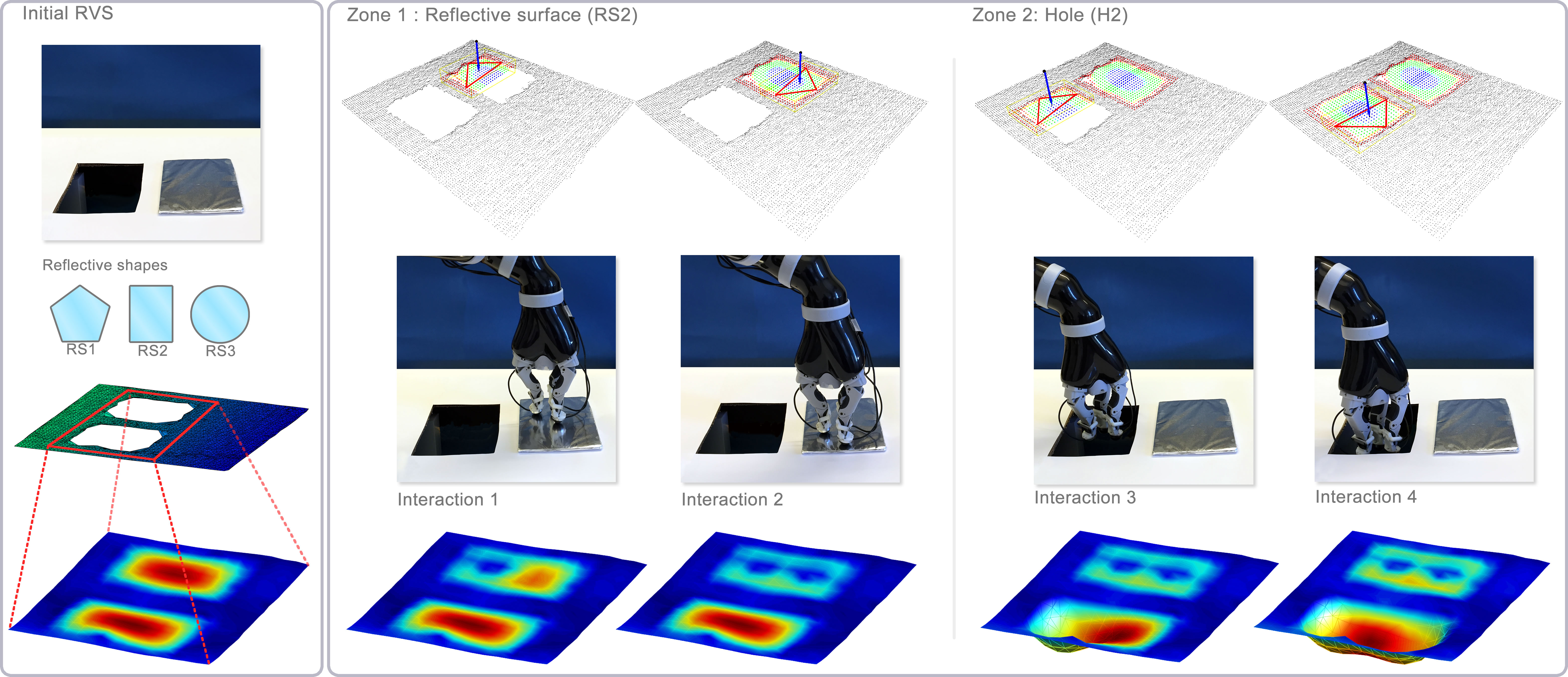}
  \caption{Exp. setup 1: Different reflective objects are placed next to holes of similar shape which generates ambiguity on the VRS point cloud. The algorithm identifies regions of high uncertainty and starts poking the surface on different locations. After 4 interaction, the 3D model allows to clearly identify the two elements. Blue color in the first row indicates high variance in the GRF model queried in proximity of the select Delaunay triangles. Red color on the third row indicates high uncertainty in the GPIS model.  }
  \label{fig:reconstr_img}
\end{figure*}

\begin{figure*}[t!]
  \center
     \includegraphics[width=\textwidth]{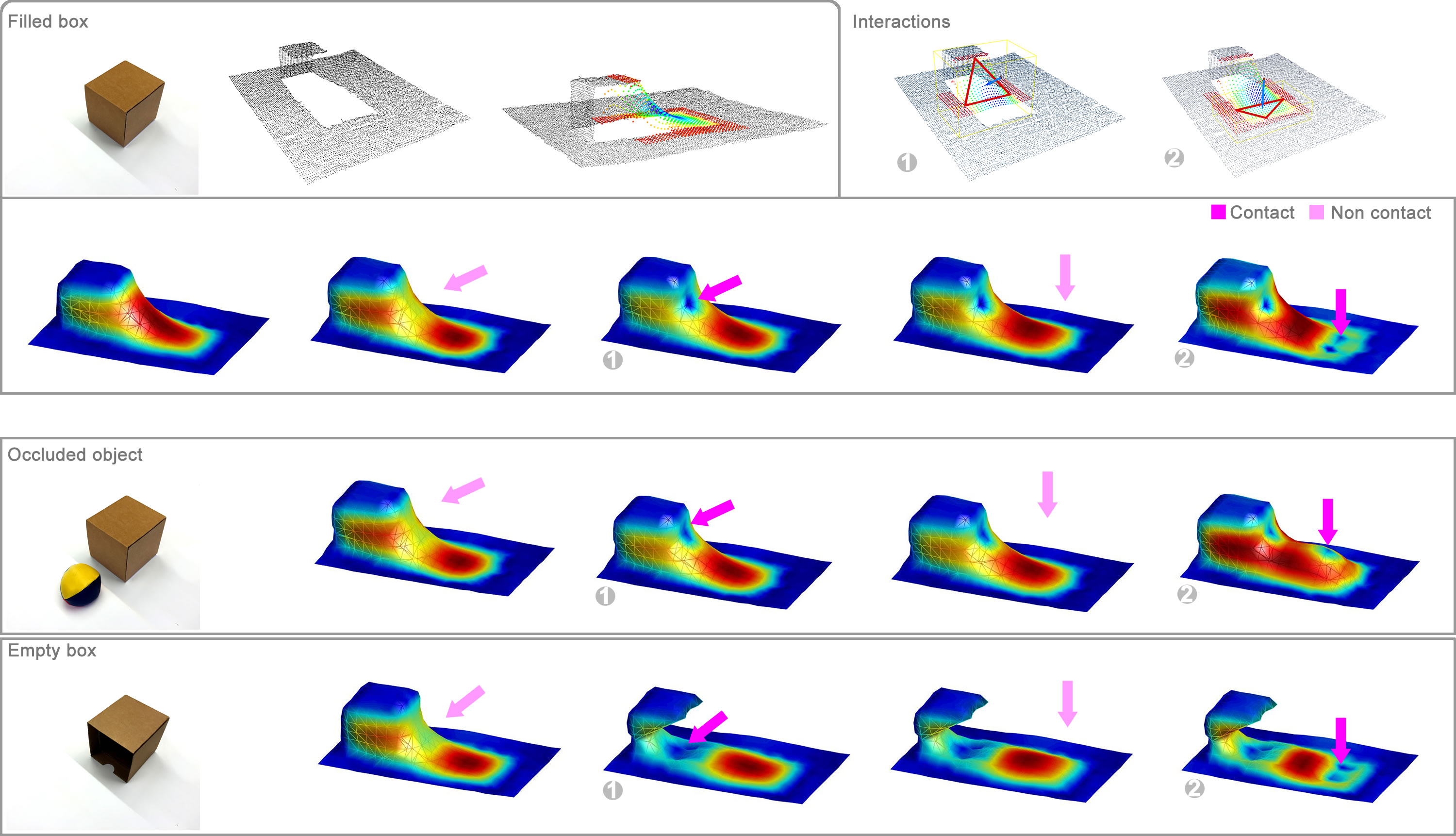}
  \caption{Exp. setup 2: An object generates an occluded area on the VRS (first row). The algorithm analyses three different situations where the occluded area is a flat terrain (second row), hides a different object (third row) or has complex shapes (fourth row). After 4 interaction, the 3D model allows to describe the occluded area. Colormaps for points (first row) and shapes are chosen as described in Fig.\ref{fig:reconstr_img}. In the last scenario the model identifies the empty area inside the box but creates an artifact (two missing faces) due to a lack in lateral interactions. }
  \label{fig:reconstr_img_box}
\end{figure*}

\begin{algorithm}
\caption{Generate a test set for a GPIS from an inferred GRF}
\label{alg:generate_dataset}
\begin{algorithmic}[1]
\Procedure{GenerateSubset}{}
\State $X_*\gets 2Dgrid(size) $
\State $f_{{rf}_*}\gets gpmean(X_*) $\Comment{from the GRF}
\State ${V}_{{rf}_*}\gets gpvar(X_*) $\Comment{return the diagonal}
\ForAll{ $\mathbf{x}$ \text{in} $X_*$} {}
		\State $y \gets f_{rf}(\mathbf{x})- \left( m * {V}_{{rf}_*}(\mathbf{x}) + \tau_v \right)$\Comment{ lower height, $m \text{ and }\tau_v$ constants}
		\While {$y < f_{rf}(x)+ \left( m * {V}_{{rf}_*}(\mathbf{x}) + \tau_v \right)$}
				\State $X_{{is}_*} \gets addNewPointToSet( \mathbf{x}, y) $
				\State $y\gets y + \Delta_y  $\Comment{$\Delta_y$ incremental constant}
		\EndWhile
\EndFor
\State \textbf{return} $X_{{is}_*}$
\EndProcedure
\end{algorithmic}

\end{algorithm}

\section{Experimental evaluation}
\label{sec:Experiments}

In the following we describe the experimental setup showed in Fig. \ref{fig:setup_arm} and the experimental scenarios.

\subsection{Experimental setup}
\label{sec:expsetup}

\begin{figure}[t]
  \center
     \includegraphics[width=\columnwidth]{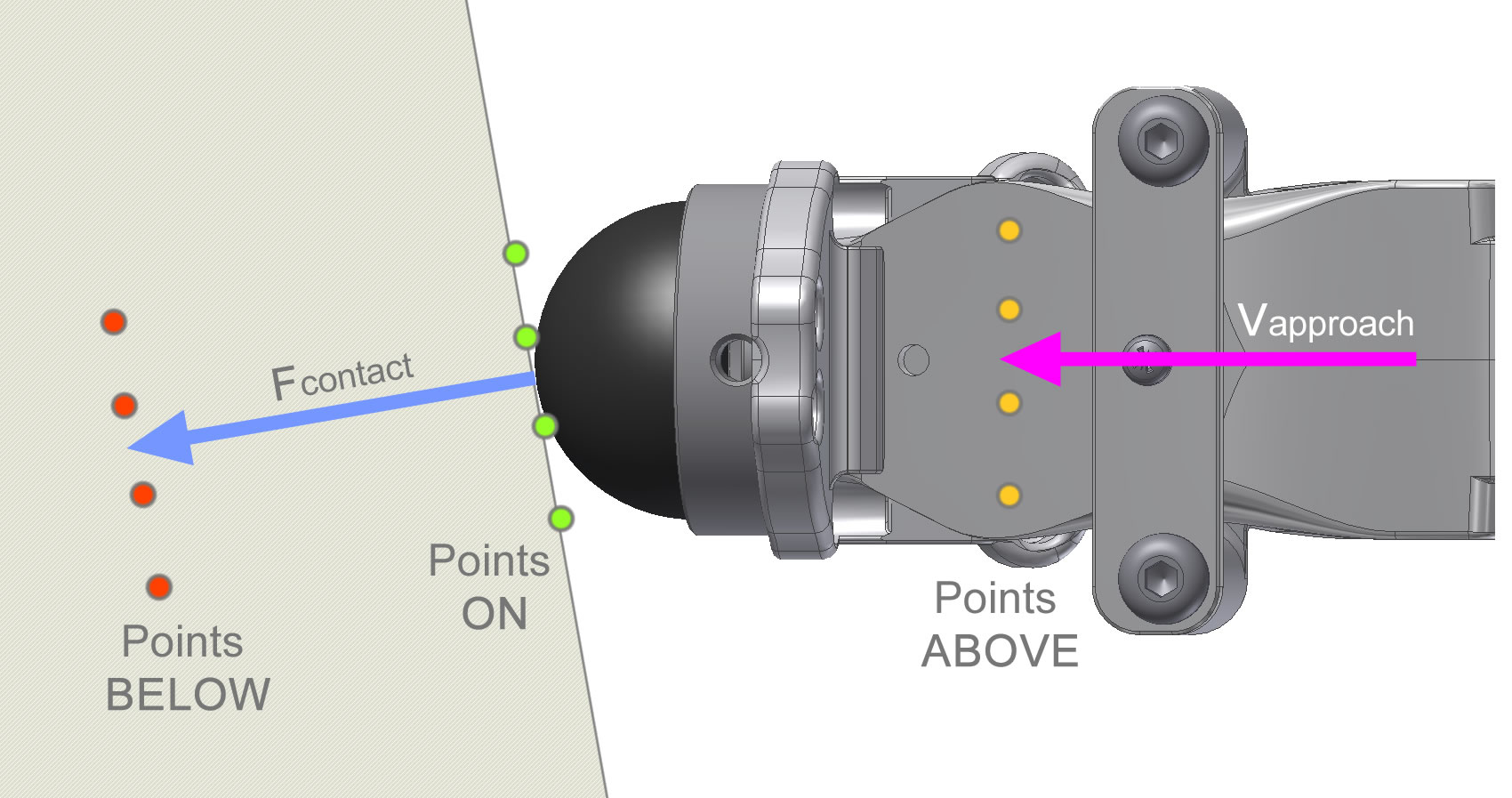}
  \caption{\small{Top view of a finger with positions of on and off surface points after contact. }}
  \label{fig:points_adding}
\end{figure}

The point cloud is obtained from a PrimeSense 3D camera placed 60 cm above a table oriented to form a $35\,^{\circ}$ angle with the table plane.
The table surface can be configured to contain holes, reflective surfaces or soft surfaces in order to recreate different scenarios.
The tactile sensory system is composed of a Kinova Jaco2\footnote{Kinova website: http://www.kinovarobotics.com/} robotic arm (6 dof) with a 3 fingered Kinova KG-3 gripper equipped with 3D OptoForce force sensors\footnote{Optoforce website: http://optoforce.com/}.
The tactile sensors can detect slipping and shear forces with high frequency.
We use the sensor output to obtain 6D force-position signals which is used for generating the tactile training dataset. Proprioceptive data are less affected by noise with respect to vision sensor data. We generate contact and non-contact points after each physical interaction using the sensor orientation w.r.t. the world frame and the output contact force.
We generate above-the-surface 3D points as square grids of 16 points placed along the downsampled fingertip trajectories (i.e. sensor position) oriented accordingly. The size of each grid is $ 8\text{mm} \times 8\text{mm}$ as the spherical sensor dimension and consistent with the VRS point cloud density. We use the contact force direction to orient the on-surface points grid to be orthogonal to the surface normal at the contact position. 
Below-the-surface points are virtually generated only in case of contact, translating a copy of the grid of on-surface points along the surface normal (see Fig. \ref{fig:points_adding}).

The arm follows the approach vector on each target triangle until contact, until the arm reaches its work space limit (i.e. VRS border) or until it diverges too much from the target (e.g. terrain holes).

Hybrid position-force control \cite{fisherhc} is used in proximity of the target point to allow small displacement along the plane orthogonal to the approach vector while imposing a minimum contact force along the approach.

\subsection{Scenarios}
\label{sec:expscenario}

\subsubsection{Reflective surfaces}
\label{sec:expincomplete}

Reflective surfaces as metal plates or water pots generate ambiguity in the point cloud representation of the environment due to photometric effects as shown in the initial VRS in Fig. \ref{fig:reconstr_img}.
In the first scenario we use the presented active perception algorithm to identify and model the difference between the two incomplete point cloud regions in an online, fully autonomous fashion.
We repeated the experiment using 3 different reflective shapes and holes (reflective shapes RS 1,2 and 3 in Fig. \ref{fig:reconstr_img}) on a single exploration step.
The first row of Fig. \ref{fig:reconstr_img} shows the selected regions of interest on the point cloud with the corresponding Delaunay triangles laying on the areas of high uncertainty (estimated from the GRF imposing a minimum triangle area) and approach vectors. Each triangle selection triggers a physical interaction (visible in the second row of each column). After each action we train the GPIS with the new tactile training previously mentioned. The last row shows the isosurface of value 0 representing the implicit surface modeled by the GPIS after each interaction. We invite the reader to note how the last two interactions allow to shape the hole but do not reduce the representation uncertainty in that region. This is because we fed the model with above-the-surface points (i.e. non contact points since the arm could not reach the bottom of the hole) that only helped identify areas where the implicit surface could not be. 

After the exploration step we used a simple threshold-based binary terrain classifier to automatically label the holes in the analyzed areas. Example of more advanced terrain classification can be seen in \cite{krebstc} and \cite{lalondetc}. Results  using 3 different reflective surfaces and hole shapes are shown on Table \ref{tab:classifier_hole} on a $0.7$m$\times 0.6$m$\times 0.6$m VRS.

\begin{table}[h]
\begin{tabular}{|l||c|c|c|c|}
\hline
 Scen. & Shape in the		& Shape in the  & $\text{n}^{\circ}$& $T_h$ Classifier \\
     		& first region & second region 		   &  interact.		  & [hole/flat/object] \\
\hline
1 & RS1 & H1  & 8 & (?,?)$\to$(flat,hole) \\
2 & RS2 & H2  & 4 & (?,?)$\to$(flat,hole) \\
3 & RS3 & H3  & 6 & (?,?)$\to$(flat,hole) \\
4 & RS1 & RS2 & 6 & (?,?)$\to$(flat,flat) \\
5 & H1  & H2  & 7 & (?,?)$\to$(hole,hole) \\
\hline
\end{tabular}
  \caption{Different combinations of reflective surface (RS) and hole (H) shapes placed in two different regions. A simple threshold classifier $T_h$ labels the presence of holes using two average heights centered on the query regions as shown in Fig.\ref{fig:reconstr_img}.  }
  \label{tab:classifier_hole}  
\end{table}  

The exploration steps, including elaboration time\footnote{Using PCL, Eigen, ROS, Kinova SDK}, planning and physical interactions lasted between 3-5 minutes for the above mentioned scenarios. The dimension of $\textit{P}_{VRS}$ varied between $N=5000$ and $N=21000$ points depending on the dimension of the occlusions and manipulabity constraint.
 $\mathbf{B_{VRS}}$ contained between $12$ and $21$ baricenter points. Dimension of $\mathbf{T_{INT}}$ ($L$, resulting after reduction based on GRF variance) was between $4$ and $8$ ($\text{n}^{\circ}$ interact.).

\subsubsection{Occluded areas}
\label{sec:expoccluded}

In the second experiment (Fig. \ref{fig:reconstr_img_box}) we demostrate how the algorithm can reconstruct occluded areas and how it can extract environmental properties which are not visible in a simple DEM. 
Similarly to the previous scenarios we analyse (now independently) elements placed inside the VRS that have the same point cloud representation as shown in Fig. \ref{fig:reconstr_img_box}. The first element is a full cubic box that generates a large occluded area on the initial VRS point cloud. The second element is an empty cubic box that hide its open face from the camera. The empty area inside the cube cannot be represented by a DEM (GRF) that would only consider the height of the upper side on the box. The third object is the same full cubic box that hides a third different object (a soft ball) placed a few centimeter behind it. By limiting the area of the Delaunay triangles we force the algorithm to only have 2 interactions on the exploration step for each scenario. 
The first row of Fig. \ref{fig:reconstr_img_box} shows the the point cloud representation of the full box on the VRS together with its uncertainty distribution generated by the GRF and the process of action selection with the selected Delaunay triangles. Second row shows the evolution of the internal representation (computed offline) during each interactions. It is possible to notice how the implicit surface changes while the arm approach the target. 
Second row shows the evolution of the GPIS model for the second scenario. The occluded object becomes visible only during the second interactions after a physical contact.
Last row shows that the GPIS can generalize information more complex than the ones embedded on a DEM. The empty cube shaped is revealed by the first interaction that do not add any on-surface point. The box is in fact carved by the off-surface points. The two triangles carrying higher uncertainty selected during the exploration step do not generate any lateral interaction with the box. This results in an artifact in the internal representation (last row of Fig.\ref{fig:reconstr_img_box}) where the two lateral faces of the box disappear. Such situation can be avoided increasing the number of interactions for each exploration step. Similarly to the previous experiment we use a simple threshold classifier on the occluded area to identify the presence of objects as shown in Table \ref{tab:classifier_box}.

\begin{table}[h]
\begin{tabular}{|l|c|c|c|}
\hline
 Scen. & Scenario description   & $\text{n}^{\circ}$& $T_h$ Classifier \\
     	  &               &  interact.		  & [object/flat/hole] \\
\hline
1 & Full box & 4   & (obj,?)$\to$(obj,flat) \\
2 & Box with occluded object  & 4 & (obj,?)$\to$(obj,obj) \\
3 & Empty box   & 4 & (obj,?)$\to$(obj,flat) \\
\hline
\end{tabular}
  \caption{ Detection of occluded objects. A simple threshold classifier $T_h$ labels the presence of objects using two average heights centered on the box and on the occluded area behind the box respectively.  }
  \label{tab:classifier_box}  
\end{table}

\section{Conclusions}
\label{sec:conclusions}

We presented an efficient probabilistic framework for building a 3D model of a surface containing different occluded areas, objects and reflective surfaces
\footnote{Video of an experiment available at: https://youtu.be/0-UlFRQT0JI }.
 The algorithm uses Delaunay triangulation and Gaussian Random Fields to quickly identify areas poorly described by the visual sensory system avoiding the computational cost of Gaussian Implicit Surfaces. The system generates subsequent target positions and orientations for a trajectory planner that brings a robotic arm equipped with tactile force sensor to touch the uncertain regions of the local map. On-surface and off-surface points generated during each physical interaction of the arm are used to update a Gaussian process implicit surface that keeps an internal complex representation of the environment. We did real experiments to show how very few interactions can unveil fundamental information hidden in the environment. We also showed how off-surface points alone (that are generated in case of non contact trajectories) can help to model simple terrain shapes. The algorithm can be used in an online process as opposed to other methods \cite{yasemingp} and can be iterated to increase the quality of the 3D model.
A limitation of the framework appears when the terrain complexity increases or when the covariance functions used for the GPIS and the GRF differ considerably. In such cases tactile interactions (that are planned using the GRF model) cannot bring enough information to the GPIS, resulting in wrong surface representations. A second weakness arises when the arm modifies too much the surface under analysis during the physical interactions and the internal representation of the environment diverges from the real world. 
In future work we plan to study these problems by segmenting objects in the environment and incorporating relative translations into the model. Variance values inside the triangles can help to generate sliding-on-surface acquisitions to obtain more tactile data from each interaction and embed additional surface properties.
We also plan to test the algorithm on the robot shown in Fig. \ref{fig:vrs_pic} and on a PR2.

\section*{Acknowledgments}
The authors gratefully acknowledge funding under the European Union's seventh framework program (FP7), under grant agreements FP7-ICT-609763 TRADR.

\bibliographystyle{IEEEtran}
\bibliography{FieldRobotics,InteractivePerception}

\end{document}